\documentclass{article}

\usepackage{arxiv}

\usepackage[utf8]{inputenc} 
\usepackage[T1]{fontenc}    
\usepackage{hyperref}       
\usepackage{url}            
\usepackage{booktabs}       
\usepackage{amsfonts}       
\usepackage{nicefrac}       
\usepackage{microtype}      
\usepackage{lipsum}
\usepackage{graphicx}
\usepackage{graphics}
\graphicspath{ {./} }

\usepackage{amsmath} 
\usepackage{amssymb}  
\usepackage{booktabs}
\usepackage{siunitx}
\usepackage{multirow}
\usepackage[dvipsnames]{xcolor}
\usepackage{hyperref}
\usepackage{tabularx}

\usepackage[export]{adjustbox}
\usepackage{algorithm}
\usepackage{algpseudocode}
\usepackage{marvosym}
\usepackage[caption=false]{subfig}

\title{Non-linear Model Predictive Control for Multi-task GPS-free Autonomous Navigation in Vineyards}

\author{Matteo~Sperti$^{1}$, Marco~Ambrosio$^{1}$, Mauro~Martini$^{1}$, Alessandro~Navone$^{1}$, Andrea~Ostuni$^{1}$ and Marcello Chiaberge$^{1}$ 
\thanks{$^{1}$ Department of Electronics and Telecommunications, Politecnico di Torino, 10129, Torino, Italy. \tt\footnotesize \{firstname.lastname\}@polito.it}}

\author{
Matteo Sperti\\
  Department of Electronics and Telecommunications \\
  Politecnico di Torino\\
  Torino, TO, 10129 \\
  \texttt{matteo.sperti@studenti.polito.it} \\
  \And
Marco Ambrosio\\
  Department of Electronics and Telecommunications \\
  Politecnico di Torino\\
  Torino, TO, 10129 \\
  \texttt{marco.ambrosio@polito.it} \\
  \And
Mauro Martini \\
  Department of Electronics and Telecommunications \\
  Politecnico di Torino\\
  Torino, TO, 10129 \\
  \texttt{mauro.martini@polito.it} \\
  \And
 Alessandro Navone \\
  Department of Electronics and Telecommunications \\
  Politecnico di Torino\\
  Torino, TO, 10129 \\
  \texttt{alessandro.navone@polito.it} \\
   \And
 Andrea Ostuni \\
  Department of Electronics and Telecommunications \\
  Politecnico di Torino\\
  Torino, TO, 10129 \\
  \texttt{andrea.ostuni@polito.it} \\
   \And
 Marcello Chiaberge \\
  Department of Electronics and Telecommunications \\
  Politecnico di Torino\\
  Torino, TO, 10129 \\
  \texttt{marcello.chiaberge@polito.it} \\
}

\begin{document}
\maketitle
\begin{abstract}
Autonomous navigation is the foundation of agricultural robots. This paper focuses on developing an advanced autonomous navigation system for a rover operating within row-based crops. A position-agnostic system is proposed to address the challenging situation when standard localization methods, like GPS, fail due to unfavorable weather or obstructed signals. This breakthrough is especially vital in densely vegetated regions, including areas covered by thick tree canopies or pergola vineyards. 
This work proposed a novel system that leverages a single RGB-D camera and a Non-linear Model Predictive Control strategy to navigate through entire rows, adapting to various crop spacing. The presented solution demonstrates versatility in handling diverse crop densities, environmental factors, and multiple navigation tasks to support agricultural activities at an extremely cost-effective implementation. Experimental validation in simulated and real vineyards underscores the system's robustness and competitiveness in both standard row traversal and target objects approach.
\end{abstract}

\section{Introduction}\label{sec:introduction}

In recent years, precision agriculture has advanced significantly, utilizing technology to optimize crop production and reduce waste \cite{zhai2020decision}. Row-based crops, in particular, represent a pivotal scenario in precision agriculture applications. Research in this domain encompasses various aspects, such as plant health monitoring \cite{monitoring}, harvesting \cite{jia2020apple}, spraying \cite{spraygrape}, irrigation \cite{irrigation}, and seeding \cite{seeding}.

This work contributes to the foundation problem of robust autonomous platforms in row-based crops \cite{cerrato2021deep, salvetti2022waypoint}, to address all the aforementioned tasks. Standard localization technologies as the Global Navigation Satellite System (GNSS), can fail in this context due to adverse weather or dense vegetation. Moreover, GPS-based solutions are often enhanced by the corrections carried out by multiple costly Real-Time Kinematics (RTK) receivers.

\begin{figure}[t]%
    \centering
    \subfloat[Straight vineyard]{
        \includegraphics[width=0.45\linewidth]{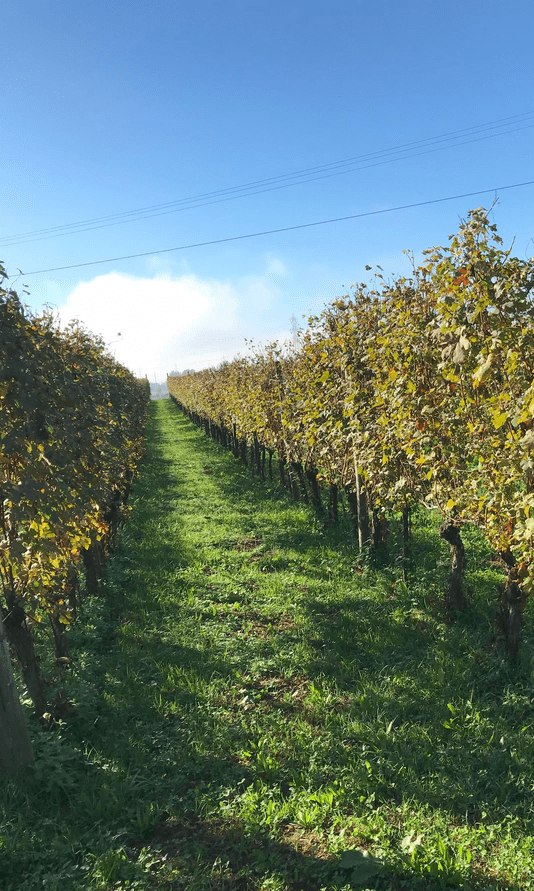}    
    }
    \subfloat[Pergola vineyard]{
        \includegraphics[width=0.45\linewidth]{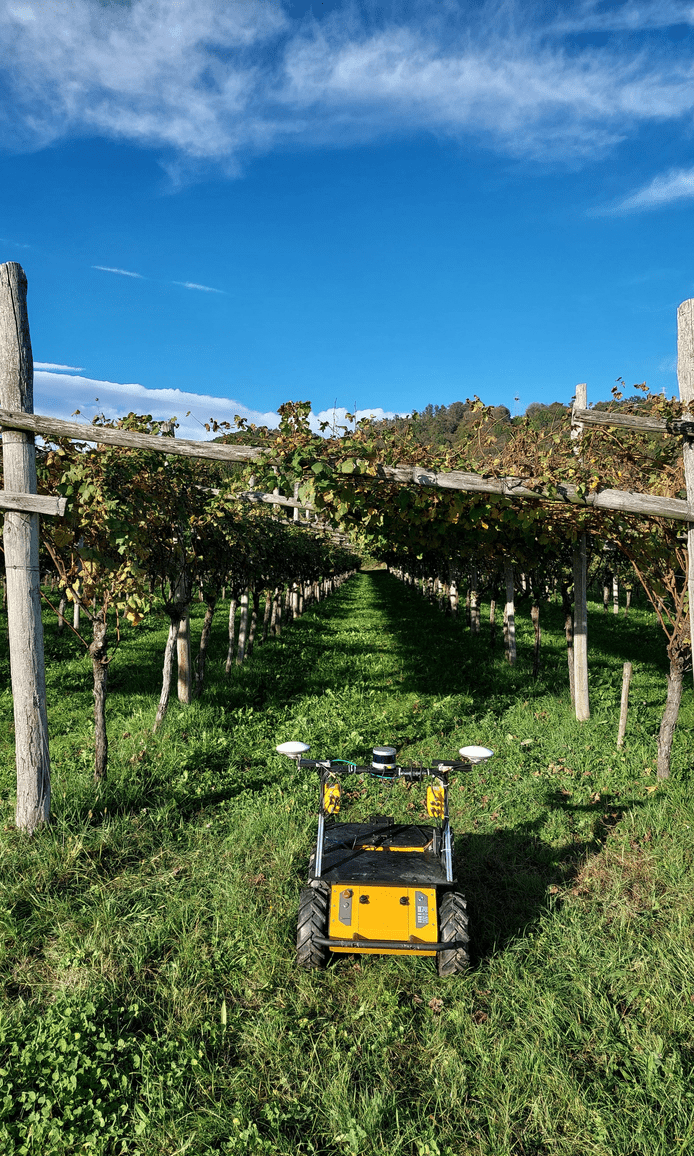}    
    }%
    \caption{Vineyards used for testing the proposed navigation system in Agliè, Turin, Italy.}
    \label{fig:real_vineyards}
    \vspace{-4mm}
\end{figure}

Alternative methods, such as Visual Odometry (VO), have been investigated to localize rovers using camera image streams \cite{zaman2019cost}. However, challenges arise in row-crop fields due to the repetitiveness of environmental visual patterns. 
A versatile position-agnostic system is therefore proposed, excelling in scenarios where traditional methods fall short. The presented control system can address multiple navigation tasks, such as traversing entire rows, avoiding obstacles effectively, and approaching target objects in varying row spacing. Position-agnostic sensorimotor agents directly map sensor data to rover velocity commands without relying on fixed Reference Frames (RFs). For instance, \cite{Aghi_2021, Navone_2023, Radcliffe_2018} proposed to segment the input image to compute a set point in the camera frame and to use a proportional controller to align the rover towards the set point. These methods, however, fail in the case of pergola vineyards or high trees in which the sky is not visible \cite{Aghi_2021, Radcliffe_2018} or the crops are not uniform on both sides \cite{Navone_2023}, see Fig. \ref{fig:real_vineyards}. Moreover, segmentation-based methods encounter difficulties in generalizing due to visual seasonal changes and in handling unexpected obstacles along the path. 

Decision algorithms provide another avenue, with Deep Reinforcement Learning agents trained by \cite{Martini_2022} for decision-making or Convolutional Neural Network (CNN) used by \cite{Huang_2021} to output actions from a discrete set. Additionally, \cite{Villemazet_2023} introduced a path-following Non-linear Model Predictive Control (NMPC) approach, leveraging a Point Cloud Data (PCD) from four cameras to generate the reference path.

\begin{figure*}[ht]
    \centering
    \includegraphics[width=0.93\linewidth]{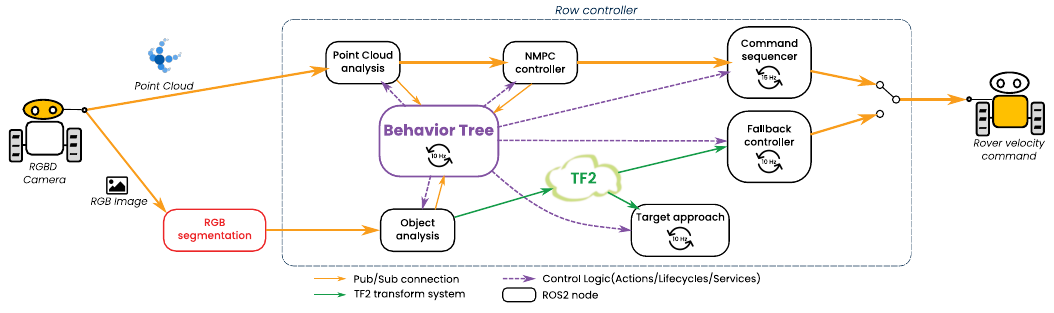}
    \caption{Computation Graph of the ROS 2 overall application system. A Behavior Tree manages and coordinates the NMPC controller for row traversal, target object approach, and recovery behaviors for robust multi-task navigation.}
    \label{fig:computation_graph}
    \vspace{-4mm}
\end{figure*}

The primary contribution of this research lies in developing a new robust controller tailored for row crop geometry, avoiding the need for precise and costly localization systems such as GPS receivers. Notably, a single RGB-D camera represents a cost-effective option compared to other sensors like 3D LIDARs.

Furthermore, the navigation system has been conceived to support task-oriented behavior. This enables the system not only to navigate the agricultural space efficiently but also to engage in auxiliary tasks: approaching target objects in the field, such as boxes or specific plants, or avoiding obstacles, and seamlessly resuming the row traversal process. This flexibility enhances the overall utility of the system and broadens its applicability for diverse agricultural tasks beyond navigation, from object transport to plant harvesting.

The next sections are organized as follows: Section \ref{sec:methodology} describes the proposed control system for multi-task position-agnostic autonomous navigation in row-based crops. Section \ref{sec:results} illustrates the experiments conducted both in simulated and real vineyards, discussing the obtained results. Finally, Section \ref{sec:conclusion} wraps up all the considerations on the study and suggests future directions.

\section{Methodology}
\label{sec:methodology}
The proposed methodology adopts a position-agnostic controller approach to guide the robot in row-based crops without relying on a localization signal. Costly GPS sensorization of the platform may lead to unreliable performances in case of thick vegetation. Taking a PCD as input, the controller computes in real-time linear velocity $v_x$ and angular velocity $\omega_z$. 

\subsection{Navigation System Architecture}

The computation graph, shown in Fig. \ref{fig:computation_graph}, illustrates the system's structure.
The overall system is orchestrated by a \textit{Behavior Tree}, overseeing high-level logic, mission switches, start and stop commands, failure detection, and initiating fallback procedures.

The RGB-D camera data is analyzed by two parallel operation flows that carry out standard row traversal and check the presence of potential objects of interest in the mission to be approached. The \textit{Point Cloud Analysis} process computes two lines that delimit the intra-row space from the PCD, necessary for the main navigation purpose. The \textit{NMPC controller} uses a Non-linear Model Predictive Control strategy to compute the control sequence given the estimated geometrical constraint of the row.
A generic \textit{Object Detection} visual algorithm could be adopted to estimate the position of potential target objects from the camera image. If any, the \textit{Target approach} process is triggered to smoothly guide the robot to a desired position near the target. A \textit{Fallback Controller} manages recovery from fault behaviors: a simple proportional controller is used to re-align the rover with the plants' row. 

\subsection{Point Cloud processing pipeline} \label{sec:vision_pipeline}
The PCD of the camera is processed to perceive potential obstacles and the boundaries of the crop row. The output of this pipeline includes the array of obstacle points and two straight lines, which represent the geometrical sides of the row.

The first part of the procedure consists of mapping the input PCD to the 2D horizontal plane. Hence, as a first thing, the PCD is transformed into the rover RF. A down-sampling, performing a voxelization operation with a resolution $r_v$, and filtering using a classical k-NN algorithm to exclude the noise points, are applied. Then, the PCD is cropped to eliminate outliers and misleading points in the sky and on the ground. Therefore, minimum and maximum height thresholds, $z_{th, min}$, $z_{th, max}$ on the z-\textit{axis} are set to ensure the removal of ground points. This operation is necessary in cases where the rover is not perfectly parallel to the ground plane due to bumpy or rough terrain. If, after this preprocessing, the fraction of remaining points, in relation to the original amount, falls below a specified threshold, $f_{points}$, the field of view is deemed empty. Consequently, no row is detected. On the other hand, if the fraction exceeds the threshold, the points are projected onto a 2D plane by considering only their coordinates on the x-y plane generating a grid map.

After the generation of the obstacle occupation map, the areas behind them are also considered occupied. This allows us to identify the inner edge of the plants in the row. Then, a heuristic approach is used to gather the occupied zones on the available borders. 

Since the two internal row borders are considered two straight lines, a least square fit is applied to evaluate the angular and bias coefficients, $a_i, b_i \in \mathbb{R}$ of the equation $y = a_i x + b_i, i \in [l,r]$. Two lines are generated, one for the left side $l$ and one for the right one $r$.

Finally, a safety distance margin $R$ is added to the row's two borders to consider the robot's occupancy and account for possible errors. Moreover, suppose the rover is required to travel only in half of the available row space, for example in a scenario where multiple robots are expected to move in opposite directions. In that case, the middle line is computed and used to separate the two motion lanes. An error is raised if one of the two lines is (given a predefined maximum angle) perpendicular to the $x-$axis, i.e., the direction of motion of the rover. Hence, the fallback recovery procedure is initiated to prevent the rover crash and realign it with the row direction.


\begin{figure}
    \centering
    \includegraphics[width=0.8\linewidth]{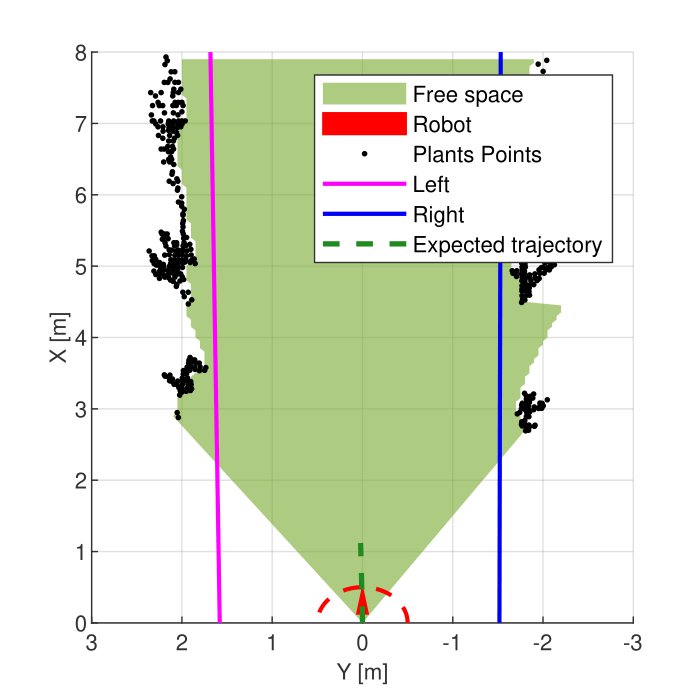}
    \caption{The black points represent the input PCD, filtered and flattened on a 2D map (obstacles), the green area is the free space in front of the rover, while the two straight lanes represent the lane borders. Finally, the dotted green line is the expected trajectory as computed by the NMPC controller.}
    \vspace{-4mm}
    \label{fig:visualization_node}
\end{figure}

\subsection{NMPC formulation}

A customized model and cost function were meticulously tailored to address the specific requirements and characteristics of the rover's navigation scenario. This involved carefully calibrating the model parameters and formulating the cost function terms, as well as the problem constraints. The inputs of the NMPC controller are the points representing the obstacles and the two first-order polynomials representing the two straight lines delimiting the lane, expressed in the robot's RF (as presented in Section \ref{sec:vision_pipeline}).

The NMPC approach requires a plant model to predict future states. For this purpose, a modified version of the Unicycle Model was selected. In particular, quaternions are used for the representation of orientation angle. Summarizing, the kinematic model of the unicycle has been modified to:

\begin{equation}
    \begin{bmatrix}
        \dot{x_1} \\
        \dot{x_2} \\
        \dot{x_3} \\
        \dot{x_4}        
    \end{bmatrix} 
     =
    \begin{bmatrix}
        v(x_3^2 - x_4^2) \\
        v(2x_3x_4) \\
        -\omega \frac{x_4}{2} \\
        \omega \frac{x_3}{2}       
    \end{bmatrix}
\end{equation}
where $x_1 = x$, $x_2 = y$, $x_3 = \cos{\frac{\theta}{2}}$, $x_4 = \sin{\frac{\theta}{2}}$.

Moreover, input saturation constraints were incorporated into the NMPC minimization problem, allowing for the specification of maximum linear and angular velocities, namely $v_{x, max}$ and $\omega_{z, max}$, as parameters before the system's initiation.

In addition, non-linear constraints were integrated to ensure obstacle avoidance, according to the following formula:
\begin{equation}
- (x_1 - o^i_1)^2 - (x_2 - o^i_2)^2 + R^2 \leq 0
\end{equation}
where the two negative terms represent the square of the Euclidean distance between the rover pose $\mathbf{x}$ and the $i$-th obstacle $\mathbf{o}^i$, and the parameter $R$ represents a predetermined safe distance between the rover and an obstacle point. This constraint must hold for each time step $t_k = 1, \dots, T_H$ and for every obstacle point, providing a robust mechanism for obstacle avoidance throughout the prediction horizon.

The core of the NMPC formulation lies in defining an objective function, which needs to be optimized, represented as follows:
\begin{equation}
   C=\sum_{k=0}^{n-1}(\underbrace{l\left(\mathbf{x}_k, \mathbf{u}_k, p\right)}_{\text {Lagrange term }}+\underbrace{\Delta \mathbf{u}_k^T \mathbf{R} \Delta \mathbf{u}_k}_{\text {r-term }})+\underbrace{m\left(\mathbf{x}_n\right)}_{\text {meyer term }} 
   \label{eq:mpc}
\end{equation}
In this equation, three contributions can be identified, respectively the $Lagrange$ term, the $meyer$ term, and the $ r$-term.

The \textit{Lagrange} term, $l\left(\mathbf{x}_k, \mathbf{u}_k, p\right)$, evaluated and summed at each time step until the prediction horizon, is composed of two contributions as in the following equation:
\begin{equation}
    l\left(\mathbf{x}_k, \mathbf{u}_k, p\right) = K_{lane} C_{lane}\left(\mathbf{x}_k, \mathbf{u}_k, p\right) +  K_{orient} C_{align}\left(\mathbf{x}_k, \mathbf{u}_k, p\right)
\end{equation}

The first term, $C_{lane}\left(\mathbf{x}_k, \mathbf{u}_k, p\right)$, aims at maintaining a central trajectory with respect to the lane while, the second term, $C_{align}\left(\mathbf{x}_k, \mathbf{u}_k, p\right)$, aims at minimizing misalignment from the row direction. The constants $K_{lane}$ and $K_{orient}$ are the weights of the corresponding contributions.

Given a position $\mathbf{x} = [x_1, x_2, x_3, x_4]$, and the two lines delimiting the row $y_l = a_lx_1 + b_l$ (on the left), and $y_r = a_rx_1 + b_r$ (on the right), the cost term regarding the lane centrality is described by the following equation:
\begin{equation}
    C_{lane} = \frac{4}{(y_l - y_r)^2} x_2^2 - 4\frac{(y_l + y_r)}{(y_l - y_r)^2} x_2 + \frac{(y_l + y_r)^2}{(y_l - y_r)^2}
\end{equation}
Essentially, it consists of a paraboloid with its minimum coinciding with the middle of the row. For each depth value $x_1$ a convex-upward parabola is constructed along the axis $x_2$ with a minimum in the middle of the lane. Therefore, the minimum cost trajectory ideally aligns perfectly with it.

The cost term for the alignment is computed considering the difference between the angular coefficient of the middle line $a_{avg} = (a_l + a_r)/2$ and the angular coefficient of a straight line oriented as the rover $a_{rover}$ as in the following equation:
\begin{align}
    a_{rover} &= \tan{\theta} = \frac{\sin{\theta}}{\cos{\theta}} = \frac{2x_3x_4}{x_3^2 - x_4^2} \\
    C_{align} &= (a_{avg} - a_{rover})^2
\end{align}

The \textit{r-}term is the quadratic penalty on changes for control inputs, which can be utilized to smooth the obtained optimal solution and serve as a crucial tuning parameter.
    
The terminal (or \textit{meyer}) term of the objective function is designed to maximize the distance traveled by the rover in the prediction horizon time interval. So, recalling that $\max{f} = \min{-f}$, the terminal (or \textit{meyer}) term is set as follows:
\begin{equation}
    m\left(\mathbf{x}\right) = - K_{travel} \frac{x_1 + a_{avg} \cdot x_2}{\sqrt{1 + a_{avg}^2}}
\end{equation}
here $K_{travel}$ represents the parameter for weighting this term, $a_{avg} = (a_l + a_r)/2$ is the angular coefficient of the line in the middle of the row, and $x_1, x_2$ are the coordinates of the rover in plane at the horizon $t_k = T_H$. The distance traveled by the rover is projected onto the middle line to weigh only the distance traveled in the direction of the row.

\section{Tests and Results}
\label{sec:results}
For testing and validation, extensive experiments were conducted on both simulated and real vineyards to illustrate the competitive advantages of the proposed solution. 

\subsection{Experimental Setting}

\begin{figure}[tb]
    \centering
    \includegraphics[width=0.8\linewidth]{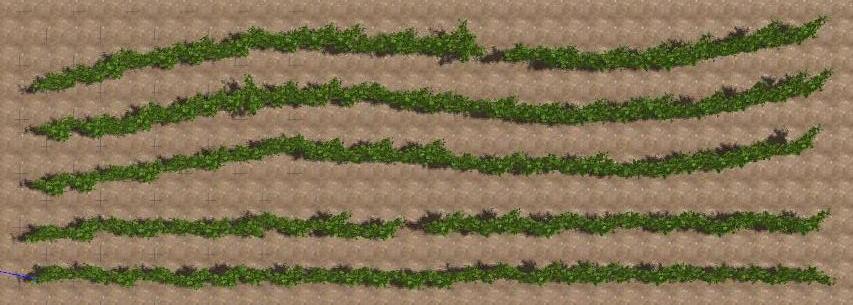}
    \caption{Aerial view of vineyards in Gazebo used for testing in simulation.}
    \label{fig:gazebo_world}
    \vspace{-4mm}
\end{figure}

All the code was developed in a ROS 2 framework and has been tested on Ubuntu 22.04 LTS using the ROS 2 Humble distro. This research employed two distinct mobile robots: the Clearpath Robotics Jackal and Husky\footnote{https://clearpathrobotics.com/
}. The first is a compact robot designed for indoor and outdoor robotics applications, while the second is a much bigger and more powerful platform. For simulated tests, the Gazebo platform, the Jackal model and description, and the  PIC4rl\_gym \cite{pic4ser_gym} evaluation tool were utilized. The world chosen, shown in Fig. \ref{fig:gazebo_world}, contains a straight and curved vineyard, with an intra-row space of around $1.5 ~m$.

Instead, tests in a real vineyard utilized the Jackal to verify the target approach feature and the Husky to evaluate the path metrics, an Intel Realsense D455 RGB-D camera, and a Velodyne VLP16 3D LIDAR for comparison. The tests were conducted on a straight vineyard with an intra-row space of around $2.5 ~m$ and on a pergola vineyard with an intra-row space of around $4 ~m$, both shown in Fig. \ref{fig:real_vineyards}.

\begin{figure}[tb]
    \centering
    \includegraphics[width=0.7\linewidth]{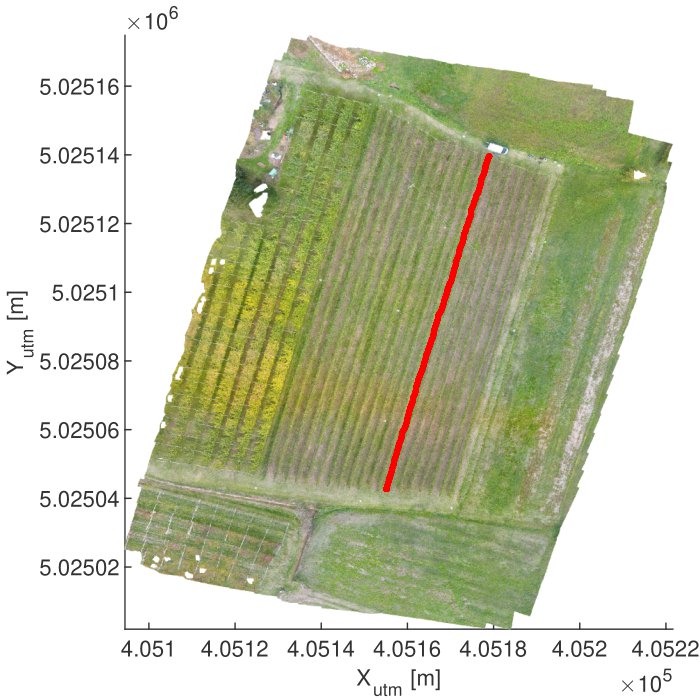}
    \caption{Satellite view of the vineyard. In red the trajectory followed by the Husky rover during a test session.}
    \label{fig:traj_gps}
    \vspace{-4mm}
\end{figure}

An accurate robot localization in the row was necessary for comparing its position to a ground truth path. However, the odometry system of the IMU of the rover failed to localize the rover due to significant drifts; SLAM techniques based on scan matching algorithms such as KISS-ICP \cite{kiss-icp}, also failed to correctly localize the system due to the repetitiveness of the environment. So, the GPS position provided by the SwiftNav Duro GNSS receiver was used as a reference to compute the metrics, along with a precise geo-localization of the row in the vineyards (Fig. \ref{fig:traj_gps}). However, GPS positioning is prone to errors in environments where leaves obstruct GPS visibility, leading to signal failures and inaccuracies in position tracking. Moreover, costs must also be considered: the GNSS receiver chosen to obtain a sufficiently precise localization is much more expensive than an RGB-D camera. These facts highlight the difficulties in localizing a ground rover in this environment and suggest the advantages of adopting a position-agnostic controller such as the one developed in this project. RGB-D cameras are also a cost-effective choice to get a limited FOV PCD, compared to a multi-range 3D LIDAR.

For the tests, $v_{x, max} = 0.4 ~m/s$ or $v_{x, max} = 0.5 ~m/s$ and $\omega_{z, max} = 0.5 ~rad/s$ has been set. The control period has been fixed to $0.7 ~s$. In the PCD processing pipeline, the resolution of the voxel has been set to $r_v = 0.05 ~m$, the minimum and maximum height threshold have been set to $z_{th, min} = 0.15 ~m$ and $z_{th,max} = 2 ~m$ and the minimum point threshold has been set to $f_{points} = 0.2$. The safety margins, for the Jackal and the Husky robots, were respectively set to $R_{Jackal} = 0.3 ~m$ and $R_{Husky} = 0.4 ~m$.

To implement the NMPC controller, the DO-MPC library \cite{do-mpc} was chosen for its versatility. The hyper-parameters of the NMPC controller have been set by a trial and error procedure.

\subsection{Evaluation Metrics}

The metrics used to evaluate the performances of the control system include:
\begin{itemize}
    \item \textit{Clearance Time} [s] and \textit{Mean linear velocity} $v_{avg}$ [m/s]: gauging the effectiveness of the proposed solution.
    \item \textit{Cumulative heading average} $Cum. \gamma_{avg}$  or \textit{standard deviation of the heading} $\gamma_{std}$ [rad], and the \textit{standard deviation of the angular velocity} $\omega_{std}$ [rad/s]: measuring the oscillation around the trajectory.
    \item \textit{Trajectory Mean Absolute Error (MAE)} [m] and \textit{trajectory Mean Squared Error (MSE)} [m\textsuperscript{2}]: measuring the error of the rover trajectory concerning a predefined ground truth, such as the center of the row or the center of the lane.
\end{itemize}

\subsection{Tests in simulated environment}

\begin{figure}[tb]
    \centering
    \includegraphics[width=0.7\linewidth]{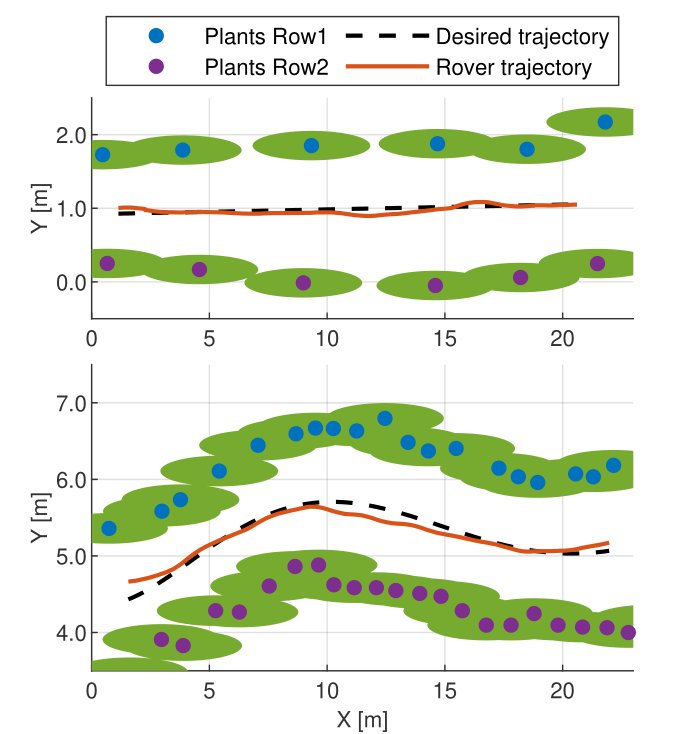}
    \caption{Tests in a simulated vineyard using the PCD of the camera as input in two different scenarios.}
    \label{fig:sim_traj}
    \vspace{-4mm}
\end{figure}

The extensive simulations conducted in simulated vineyard environments have demonstrated the reliability and robustness of the proposed navigation system. As illustrated in Fig. \ref{fig:sim_traj}, the rover's trajectory closely aligns with the desired central path, exhibiting minimal oscillations in both straight and curved vineyards.

\begin{table*}[ht]
\centering
\caption{Results of conducted experiments in simulated straight and curved vineyards.}
\label{tab:simulated_tests}
\resizebox{.9\textwidth}{!}{%
\begin{tabular}{@{}ll|cccccc@{}}
\toprule
  \textbf{Field} &
  \textbf{Sensor} &
  \textbf{Clearance time {[}s{]}} &
  \textbf{Cum. $\gamma_{avg}$ {[}rad{]}} &
  \textbf{$v_{avg}$ {[}m/s{]}} &
  \textbf{$\omega_{std}$ {[}rad/s{]}} &
  \textbf{MAE {[}m{]}} &
  \textbf{MSE {[}m\textsuperscript{2}{]}} \\ 
\midrule
 & LIDAR & 49.528±0.167 & 0.036±0.001 & \textbf{0.395±0.002} & \textbf{0.034±0.001} & \textbf{0.034±0.001} & \textbf{0.001±0.000} \\
  
 & PCD cam & 52.586±4.130 & 0.045±0.001 & 0.377±0.019 & 0.038±0.001 &  0.048±0.005 & 0.003±0.001 \\
 
 \multirow{-3}{*}{\textbf{Straight}} & RGB-D cam & \textbf{49.321±0.356} & \textbf{0.011±0.005} & 0.395±0.001 & 0.046±0.004 & 0.104±0.011 & 0.018±0.004 \\

\midrule

 & LIDAR & 52.080±0.220 & -0.024±0.001 & \textbf{0.397±0.001} & \textbf{0.036±0.001} &0.102±0.001 & 0.015±0.000 \\
 
 & PCD cam & 52.157±0.673 & \textbf{0.002±0.002} & 0.393±0.002 & 0.041±0.003 & \textbf{0.068±0.004} & \textbf{0.007±0.001} \\

 \multirow{-3}{*}{\textbf{Curved}} & RGB-D cam & \textbf{51.763±0.228} & -0.011±0.002 & 0.394±0.001 & 0.056±0.007 & 0.188±0.005 & 0.051±0.003 \\

\bottomrule
\end{tabular}%
}
\end{table*}

Detailed results are provided in Tab. \ref{tab:simulated_tests}, revealing several key performance indicators. In both straight and curved vineyards, the rover consistently achieves speeds close to the maximum limit ($v_{avg} \simeq 0.39 ~m/s$ for $v_{x, max} = 0.4 ~m/s$), resulting in effective clearance times. The rover's trajectory shows minimal oscillations, as indicated by a small standard deviation of angular velocity ($\omega_{std} \simeq 0.05 ~rad/s$), reflecting stable and smooth behavior. Path metrics, including MAE and MSE, are minimal, on the order of centimeters. This demonstrates the rover's precise adherence to the center of its lane. In the curved vineyard, a slightly larger path error is observed (MAE up to $0.2~m$ in the worst case), attributed to the controller's inclination to cut curves. This behavior can be mitigated through parameter tuning. The algorithm's consistent performance across input sensors, including RGB-D cameras, highlights its reliability and versatility. This robustness, even compared to more expensive technologies such as LIDAR, underscores the algorithm's adaptability to various sensor configurations. The ability to achieve comparable results with RGB-D cameras suggests a cost-effective alternative for applications where LIDAR may be cost-prohibitive.
Overall, these findings underscore the effectiveness and versatility of the proposed navigation system across diverse vineyard scenarios.

\subsection{Tests in real scenario}
\begin{figure}[t]
    \centering
    \includegraphics[width=0.7\linewidth]{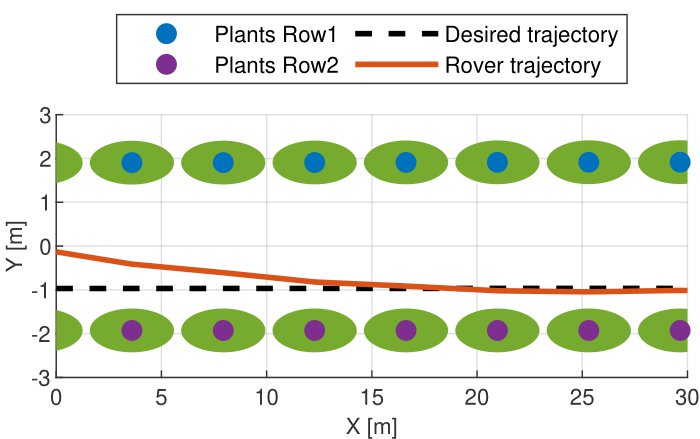}
    \caption{Test in a real pergola vineyard using the PCD of the camera as input. The desired position is in the middle of the right lane (so at 3/4 of the entire intra-row space). The rover starts in the middle of the row and then converges smoothly to the desired position.}
    \label{fig:real_pergola}
    \vspace{-4mm}
\end{figure}

The real-world tests conducted in vineyards have validated the results obtained in the simulated environment. Detailed results are presented in Tab. \ref{tab:real_tests}, highlighting the robust performance of the controller in real scenarios. As in simulation, the rover consistently achieves speeds close to the maximum limit ($v_{avg} \simeq 0.399 ~m/s$ for $v_{x, max} = 0.4 ~m/s$ and $v_{avg} \simeq 0.49 ~m/s$ for $v_{x, max} = 0.5 ~m/s$). The achieved trajectory shows minimal oscillations, as indicated by a small standard deviation of angular velocity ($\omega_{std} \simeq 0.05 ~rad/s$), reflecting stable and smooth behavior. The exception is the narrow straight vineyard in the right lane configuration, where this metric is slightly larger ($\omega_{std} \simeq 0.18 ~rad/s$): the rover displays a more oscillatory behavior, likely due to the proximity of the right lane to the crops. This behavior is less prominent in the pergola vineyard test (Fig. \ref{fig:real_pergola}) with a larger intra-row distance ($4 ~m$), where the rover shows a smooth convergence to the right lane without significant oscillations. Path metrics, including MAE and MSE, are minimal, on the order of centimeters (up to $20 ~cm$ for the narrow vineyard and up to $30 ~cm$ for the larger pergola vineyard). However, it's important to consider the error in the reference trajectory, as well as the error of localization, affected by the intrinsic accuracy of the sensor used when interpreting these results.

Moreover, using a simple recognition system for fruit boxes, the smaller robot achieved to approach the desired object of interest without any collision.

\begin{table*}[tb]
\centering
\caption{Results of a series of experiments in real vineyards.}
\label{tab:real_tests}
\resizebox{.9\textwidth}{!}{%
\begin{tabular}{@{}llcc|ccccc@{}}
\toprule
  \textbf{Field} &
  \textbf{Sensor} &
  \textbf{Position} &
  \textbf{$v_{x, max}$ {[}m/s{]}} &
  \textbf{$\gamma_{std}$ {[}rad{]}} &
  \textbf{$v_{avg}$ {[}m/s{]}} &
  \textbf{$\omega_{std}$ {[}rad/s{]}} &
  \textbf{MAE {[}m{]}} &
  \textbf{MSE {[}m\textsuperscript{2}{]}} \\ \midrule
               & & Centered   & 0.4 & 0.031±0.007 & 0.399±0.000 & 0.042±0.002 & 0.165±0.007 & 0.035±0.000 \\
 & \multirow{-2}{*}{PCD camera} &
  Right lane &
  0.5 &
  0.388±0.395 &
  0.488±0.007 &
  0.184±0.108 &
  0.204±0.098 &
  0.070±0.044 \\
\multirow{-3}{*}{\textbf{Straight}} & LIDAR & Right lane & 0.5 & 0.0153      & 0.4989      & 0.0271      & 0.1519      & 0.0294      \\

\midrule

 & &
  Centered &
  0.4 &
  0.122 &
  0.399 &
  0.063 &
  0.313 &
  0.129 \\
\multirow{-2}{*}{\textbf{Pergola}} & \multirow{-2}{*}{PCD camera} &
  Right lane &
  0.4 &
  0.047 &
  0.399 &
  0.04 &
  0.092 &
  0.011 \\

\bottomrule
\end{tabular}%
}
\end{table*}

\section{Conclusions}
\label{sec:conclusion}
The position-agnostic NMPC controller proposed in this paper has demonstrated robustness in effectively handling the diverse challenges presented in traversing row-based fields with different characteristics without accessing any localization information. Its resilient navigation on rough terrains underscores its adaptability to real-world agricultural conditions with a lower platform cost. This research significantly contributes to the continuous advancement of precision agriculture and the evolution of autonomous navigation systems tailored for row-based crop environments.

\subsection*{Acknowledgements} This work has been developed with the contribution of Politecnico di Torino Interdepartmental Center for Service Robotics PIC4SeR \footnote{\url{www.pic4ser.polito.it}}.

\bibliographystyle{unsrt}  
\bibliography{bibliography}  


\end{document}